\documentclass{svproc}
\usepackage{url}
\usepackage{graphicx}

\usepackage[backend=biber,style=iee]{biblatex}
\usepackage{textcomp}
\usepackage{hyperref}
\begin{document}
\mainmatter              
\title{Bots and Blocks: Presenting a project-based approach for robotics education}
\titlerunning{Bots and Blocks: project-based education}  
%
\author{Tobias Geger, Dominique Briechle \and Andreas Rausch}
\authorrunning{Tobias Geger et al.} 
\tocauthor{Tobias Geger, Dominique Briechle and Andreas Rausch}
\institute{Clausthal University of Technology, 38678 Clausthal-Zellerfeld, Germany,\\
\email{\{thomas.tobias.marcello.geger; dominique.fabio.briechle; andreas.rausch\}@tu-clausthal.de}}

\maketitle              

\begin{abstract}
To prepare students for upcoming trends and challenges, it is important to teach them about the helpful and important aspects of modern technologies, such as robotics. However, classic study programs often fail to prepare students for working in the industry because of the lack of practical experience, caused by solely theoretical lecturing. The challenge is to teach both practical and theoretical skills interactively to improve the students' learning. In the scope of the paper, a project-based learning approach is proposed, where students are taught in an agile, semester-spanning project how to work with robots. This project is part of the applied computer science degree study program Digital Technologies. The paper presents the framework as well as an exemplary project featuring the development of a disassembly software ecosystem for hardware robots. In the project, the students are taught the programming of robots with the help of the Robot Operating System (ROS). To ensure the base qualifications, the students are taught in so-called schools, an interactive mix of lectures and exercises. At the beginning of the course, the basics of the technologies are covered, while the students work more and more in their team with the robot on a specific use case. The use case here is to automate the disassembly of build block assemblies.
\end{abstract}

\section{Introduction} 
The usage of robotics in industry is increasing from year to year, as the World Robotic Report of 2025 is showing \cite{IFR2025WorldRobotics}.
The integration of robots is therefore no longer just focused on classical applications in the industrial sector, but is also finding its way into other sectors such as medical care, gastronomy, and logistics. 
Dhanda et al. \cite{DHANDA2025102937} described the use of Robots in general to support humans in their daily tasks and take over tasks that are both repetitive and physically challenging.
The ability to work with robots is, therefore, becoming increasingly important for contributing to this development. 

Robotics is, however, a wide-spanning knowledge field consisting of several domains, which have to be taken into account. 
To this end, it is essential to learn about programming, sensor technology, system integration, data evaluation, and functional safety, not only to operate robot-related processes but also to shape them actively. Although the theoretical knowledge in these fields is highly important, the applicability is fostered by actually using these skills. In the industry, there is often no time to build the application knowledge for all these necessary competencies in detail. Ideally, these skills are, therefore, already acquired during the study program. However, in traditional study program curricula, there is hardly any time left to integrate application phases for using and refining those skills. The benefit of introducing practical experiences already in the study program helps students to consolidate their knowledge, as \cite{schwichow2016students} explains. 

The ideal study program is, therefore, combining both theoretical knowledge and practical experiences. In addition, new support technologies for programming and development, like (LLM)-based tools, offer new possibilities for students but at the same time new challenges for lecturers to ensure their comprehension of the material. Although this shift in methodology is already visible in some lecture programs, the lasting integration of such methods requires a fundamental change in today's curricula.
New concepts are needed to explain robotics to students and give them an interactive and meaningful experience in handling robots.
Therefore, long-term support for the students should exist, which is not sustainable in a single course.
A multi-semester program is needed, where the amount of needed skills in a variety of robotics is tackled, and application modules allow students to acquire first-hand experiences. This also allows the refinement of skills over the time of study.
The paper presents the Digital Technologies project framework as an integral part of the study program and highlights its application, showing an example project featuring the development of a robot software ecosystem for an automated disassembly system.

The paper is structured as follows:
The \autoref{SotA} will deal with the state of the art, covering current educational approaches and assessments.
In \autoref{Educational Approach}, the foundational teaching approach for working with robots and in computer science in general is presented, as well as the agile project framework used in the same.
The \autoref{projectintro} contains the application scenario and the teaching methodology, where the example use case will be introduced, and the relevance in relation to the circular economy is mentioned to highlight the necessity of the approach.
The robotic system is introduced, and the development until reaching the defined project goals of the scenario is presented.
The \autoref{conclusion} summarizes the results and presents a future outlook.

\section{State of the art} \label{SotA} 
Robotics in teaching has been an integral part of automation-related and computer science colloquia around the world for quite some time. Robotics in education has, for example, been tracked in the Web of Science (WoS) since 1975 \cite{electronics10030291}. López-Belmonte et al. \cite{electronics10030291} have investigated the production of documents in different time frames and have analyzed that publications in the field of robotics in education have been drastically increasing in the last few years.

Not only has robotics played a significant role as a dedicated training field in the past, but it has also been used for motivating young students to become interested in computer science. Examples are, however, the findings of Inse and Koc \cite{https://doi.org/10.1002/cae.22321} on the skill development of students participating in the Young Engineers' Workshop (YEW). Thereby, students could increase their personal abilities in algorithmic and critical thinking related to computational thinking (CT) as well as robotics-related project development, just to name a few. Similar results in terms of improvement of CT skill were presented by Qu and Fok \cite{Qu2022}, drawing a correlation between the student-robot interaction and the enhancement of the abilities. The indication of such an improvement was stated as well by Atmatzidou and Demetriadis in 2014 \cite{Atmatzidou2014SupportCT}, especially considering their abilities regarding Educational Robotics.

In terms of a suitable framework for integrating Educational Robotics in current teaching curricula, Catlin et al. \cite{catlin2015edurobotics} have introduced both frameworks for the integration of research findings as well as the application of the same by teachers around the world. The application of the frameworks is shown by an example turtle bot called Roamer. Ideas for the curricular integration of robotics-related study fields are presented by Rubenzer et al. \cite{ready2015}, spanning from primary school to high school, thereby offering a consecutive curriculum helping the students refine their skills over substantial topics. Swenson \cite{Swenson2015robotics} has investigated the effectiveness of such programs by interviewing senior students regarding their capabilities to deal with more complex problems, indicating the improving abilities of consecutive courses. Robotic education has even been conducted in a decentralized manner as described by Yu and Silva \cite{YU2021482}, highlighting the opportunity of flexibility in robotics-related workshop and project design. 
The curricular benefit of combining theoretical education with hands-on project-based experience has also been investigated by Yilmaz et al. in 2012 \cite{Yilmaz2012}.

In terms of the implementation of flexible and agile project-based frameworks, eduScrum\textsuperscript{\textregistered} has become one of the main contenders in software development-near domains for educational purposes. eduScrum\textsuperscript{\textregistered} is based on the original Scrum framework  developed by Schwaber and Sutherland \cite{schwaber_sutherland_2020}. In contrast to the original framework, defined in the Scrum Guide \cite{schwaber_sutherland_2020}, eduScrum\textsuperscript{\textregistered} \cite{eduscrum_how_works_2025} adjusted the roles and the project steps with emphasis on teaching and accompanied learning to guide students during their assignments \cite{edu_scrum_guides_2025}. The successful implementation of eduScrum\textsuperscript{\textregistered} in a curriculum has been evaluated by Neumann and Baumann, who investigated the impact of integrating real-world problems as projects in a Master course \cite{9637344}. Similar conclusions were drawn by Kuz as well as Fernandes et al. during the application of eduScrum\textsuperscript{\textregistered} to highlight the applicability for educational purposes \cite{kuz2021scrum}\cite{educsci11080444}. In addition to eduScrum\textsuperscript{\textregistered}, Scrum Higher Education is another adaptation of the original Scrum framework with a focus on educational application and the impact of agile project management on learning \cite{9125304}. Preliminary to the project setup presented in this paper, the authors have experienced as well the benefits of agile project conduction of a Scrum adaptation, presented in the context of the development of a basic automated disassembly system \cite{10949991}.

\section{Educational Framework of the Digital Technologies study program} \label{Educational Approach} 
As pointed out in the \autoref{SotA}, project-based learning is a highly successful way for students to acquire practical experience and prepare for their professional career. This spirit is also the foundation of the study program Digital Technologies, which was established in 2019 \cite{stenzel2025projektbasierte} as a consecutive bachelor's and master's program. The study program curriculum consists one one-third of project-based courses.
In these courses, students work in small teams, typically consisting of four to six students, to develop a solution for an industry-relevant problem together. Since the preliminary courses are mainly part of the bachelor's program, the authors are focusing on the undergraduate program for the remainder of the paper. 

The bachelor’s degree program in Digital Technologies is a practice-oriented computer science program that focuses on digitalization and sustainability. It combines foundations in mathematics, engineering, and computer science with application-specific knowledge and interdisciplinary, project-based courses.
The bachelor's degree consists of 180 European Credit Transfer and Accumulation System (ECTS) points, of which 60 ECTS points are project-based courses. Each of these project-based courses is worth ten ECTS points.
In the first year, the students work on two introductory projects within their own cohort. In the following projects, the students are mixed with students from higher bachelor's degree levels or even master's students, which allows them to collaborate in more complex, interdisciplinary projects and benefit from the experience of the more advanced students, as mentioned in \cite{stenzel2025projektbasierte}.

As stated, the projects in the curriculum of the consecutive study program Digital Technologies are organized in an agile manner. 
One of the main approaches here is the Scrum Framework, which is widely used in industry \cite{MARNADA2022290}, especially in the field of software engineering and development.
The Scrum framework, at its core, allows the self-management of development teams by distributing different roles and responsibilities to the team members. Adapted to an educational environment, these roles include:
\begin{itemize}
    \item the \textit{Product Owner (PO)}, responsible for addressing the stakeholders' interests and, therefore, responsible for ensuring a satisfactory outcome and represented by the scientific staff.
    \item the \textit{Scrum Master (SM)}, responsible for coaching the team, ensuring the compliance of the defined Scrum rules, and preparing the backlog management. The position of the SM is thereby filled out by a senior student. 
    \item the \textit{Development Team (DT)}, responsible for developing and supporting the product generation. The DT consists of all the students in a team. 
\end{itemize}

\begin{figure}
    \centering
    \includegraphics[width=1.0\linewidth]{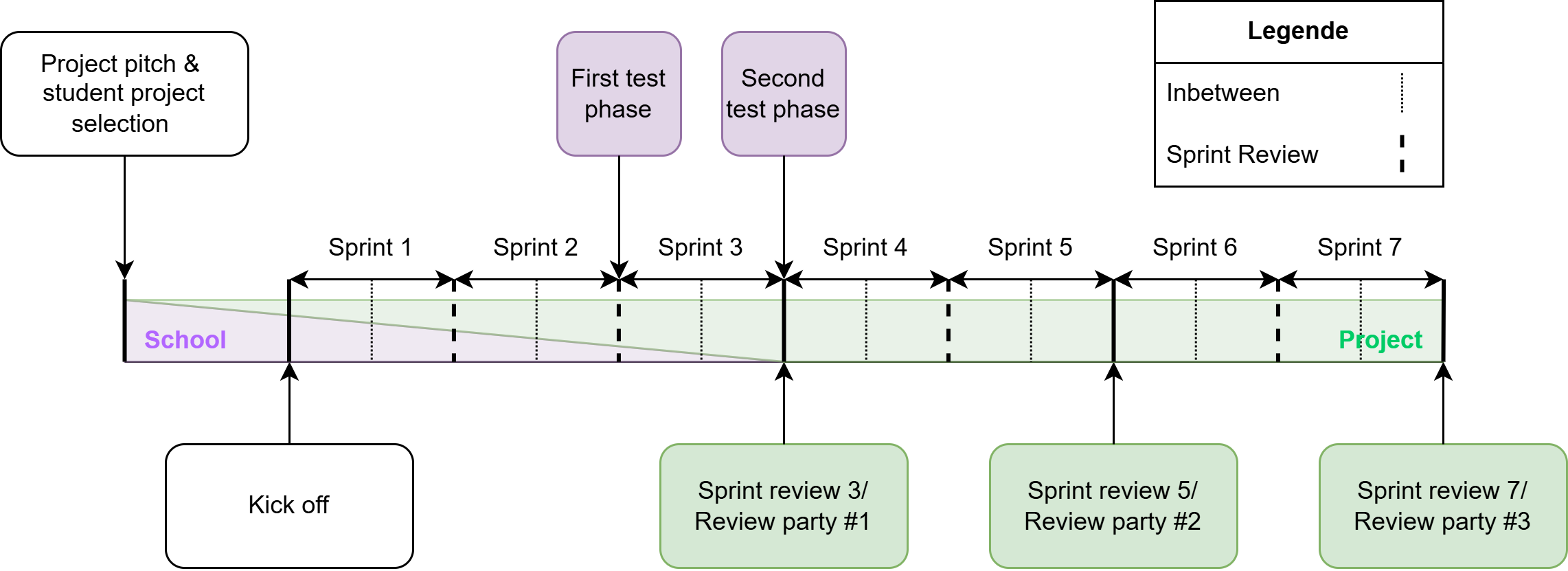}
    \caption{Project Timeline and Phase overview}
    \label{fig:Scrum}
\end{figure}

In contrast to classic project management approaches, the project team is not driven by consecutive tasks to develop a product but by the development of a minimum viable product (MVP), which is characterized by requirements stated in the user stories of the PO. As shown in \autoref{fig:Scrum}, a Scrum project is structured by so-called \textit{Sprints}, which feature a duration of two weeks and mark incremental time frames in which the DT is working on the selected user stories. 
The Sprints, in turn, are structured by Daily's, which take place twice a week, and Inbetweens, an intermediate meeting after a week in an ongoing Sprint. During the duration of the project, the DT is presenting their intermediate results at the \textit{Review Parties}. The \textit{Review Parties} are centralized events in the scope of the Digital Technologies study program, where the students present the status of their project to one another and exchange ideas at the affiliated exhibition. The third \textit{Review Party} marks the end of the project and is at the same time the date for the final presentation of the results. 

The focus of the projects over the different semesters is organized in a consecutive way.
Therefore, the project of the first semester focuses on teaching project-based skills, especially SCRUM, whereas the second semester project additionally trains the students as full-stack developers and prepares them for the projects of the following semesters. This is traditionally done by developing a browser-based computer game based on an already existing board game.

Regarding the project presented in the following, we developed a coupling of theory and practical learning in so-called schools to enable students to apply useful tools like REST, TypeScript, and the crucial robotics project tool, the Robot Operating System (ROS). In the school, the central concepts are introduced, demonstrated, and then tried out by the students. The students present, reflect, and discuss their results in the plenum. Each school covers a different topic and always begins with a short lecture of around sixty minutes. After a live demonstration by a supervisor, students work on a matching exercise while instructors support them in the meantime. Results are presented and discussed collectively, with critical parts addressed explicitly. In some cases, the students also have to solve tasks afterwards, which will then be the base for the next school session to continue with related topics or part of an intermediate exam.

The schools are embedded in the project timeline. The first two weeks are for onboarding the students and focusing on the methodological foundation, while the projects start in parallel. From weeks three to six, the time spent in schools decreases steadily as independent project work gains importance. From week seven onward, students focus primarily on their project. Milestones were the understanding of existing ROS code, publishing ROS topics, and controlling simple data flows by the end of week two. Building on this, students implemented their own ROS nodes, defined topics, services, and actions, and modeled the system’s ROS architecture. Therefore, not only code should be produced, but also design and architecture become important as well.

Assessment was conducted through two intermediate exams. The first, after week five, consisted of presenting a ROS-based solution and answering technical questions. The focus lay on the students to understand their solution and avoid the blind use of AI tools like chatbots. The second intermediate exam covered an extended use of ROS combined with a trained AI for object detection.
While we allowed the usage of AI tools like ChatGPT und CoPilot, we wanted to ensure the students understood their developed solution. The intermediate exams tackled this situation by forcing students to dive deep into their code and convince the examiners with their knowledge.

The two major contents of the schools and the project were ROS and the usage of AI for object detection and localization. ROS is an open-source framework for distributed robotic systems. The concept in detail is described in \cite{quigley2015programming}. Here, a short overview will be presented. It provides standardized components for sensors, planning, controlling, and integration of complex hardware systems. The architecture consists of processes, represented as nodes. These nodes communicate via topics, a publish-subscribe mechanism, services for request and reply, and actions for long-running tasks. There also exist interfaces, which are defined as message, service, or action type. ROS is coded in C++ or Python. The development of ROS started in 2007, and in 2015, the first alpha of ROS2 was released.

During the ROS schools, students were introduced to core terminology and tooling for interacting and developing with ROS. A live demo showed a minimal working example while covering the basics. The students had to reproduce these examples and adapt them slightly. For that, the example of the TurtleBot was used as a foundation. The visualization helped to gain better insight. In the following step, they developed their own ROS nodes and interfaces, which included creating packages, defining messages and services, using launch files, and handling the node lifecycle. Students implemented a new node that provides an additional topic and integrated it into the overall system.

As already mentioned, we taught the students about an architectural perspective while using ROS, because modeling a ROS system differs from classical, already known systems. The focus is on showing the interaction of the nodes via their topics and services using the ROS interfaces.
Also, methods for the application of robotics were shown.
The students were introduced to a hierarchical control system and how to model it.
This increased the understanding of the  ROS system as a whole by a multilevel architecture, which is needed to represent the complex variety of the system.

The second school was the school on object detection and localization. After a theoretical introduction to data preparation, labeling, training pipelines, and evaluation metrics, students applied object detection via the You Only Look Once (YOLO) algorithm. Therefore, they labeled a small dataset, ran a training job, and interpreted the results in plenum. 

The first two weeks have been exclusively reserved for working and attending the schools. The expectation was that the students could understand and publish ROS topics. Then the focus shifted, beginning from the third week and ending at week six, towards the actual project tasks. In week five, the first intermediate exam was held with a focus on the presentation of the ROS code and architecture while answering questions about their solution. After completing all schools, a second intermediate exam was conducted. From week seven onward, students work exclusively on the project and apply the knowledge from school to the project.

\section{Framework application \& project conduction} \label{projectintro} 

To highlight the applicability of the educational approach presented in \autoref{Educational Approach} for robot-based projects, we present here the findings of conducting a robot-based project in the winter semester 2025/2026 class. The projects in the study program are, as mentioned in \autoref{Educational Approach}, linked to real-world issues and are, therefore, contributing to an actual problem. The project managed by the authors as POs and conducted by the student team focuses on the automated disassembly of products in the context of the Circular Economy to salvage valuable parts, which can be reused for repair and remanufacture. Since actual products often feature more complex parts, the project goal is transferred to a more application and testing-friendly product in the scope of a student project, which is, in our case, plastic building bricks. 

\begin{figure}[htbp]
    \centering
    \begin{minipage}[t]{0.25\linewidth}
        \centering
        \includegraphics[width=\linewidth]{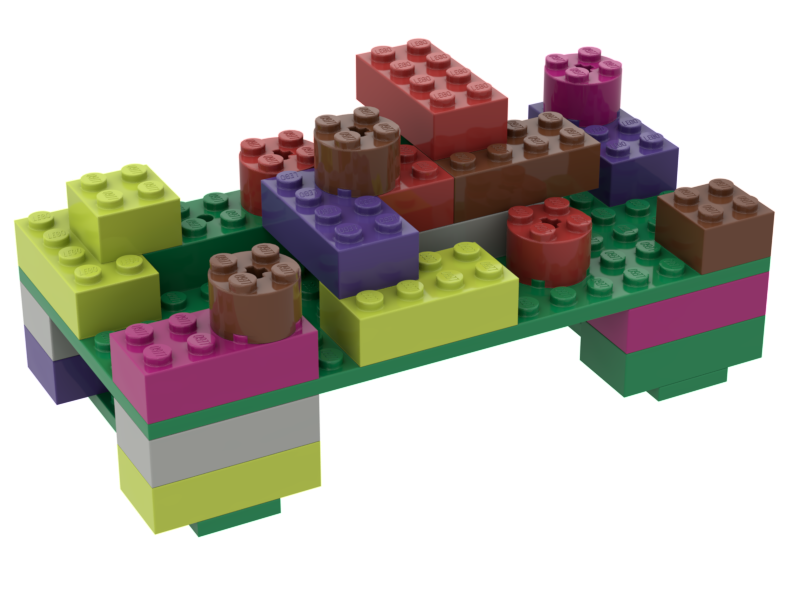}
        \caption{Example}
        \label{fig:brick_construction}
    \end{minipage}\hfill
    \begin{minipage}[t]{0.75\linewidth}
        \centering
        \includegraphics[width=\linewidth]{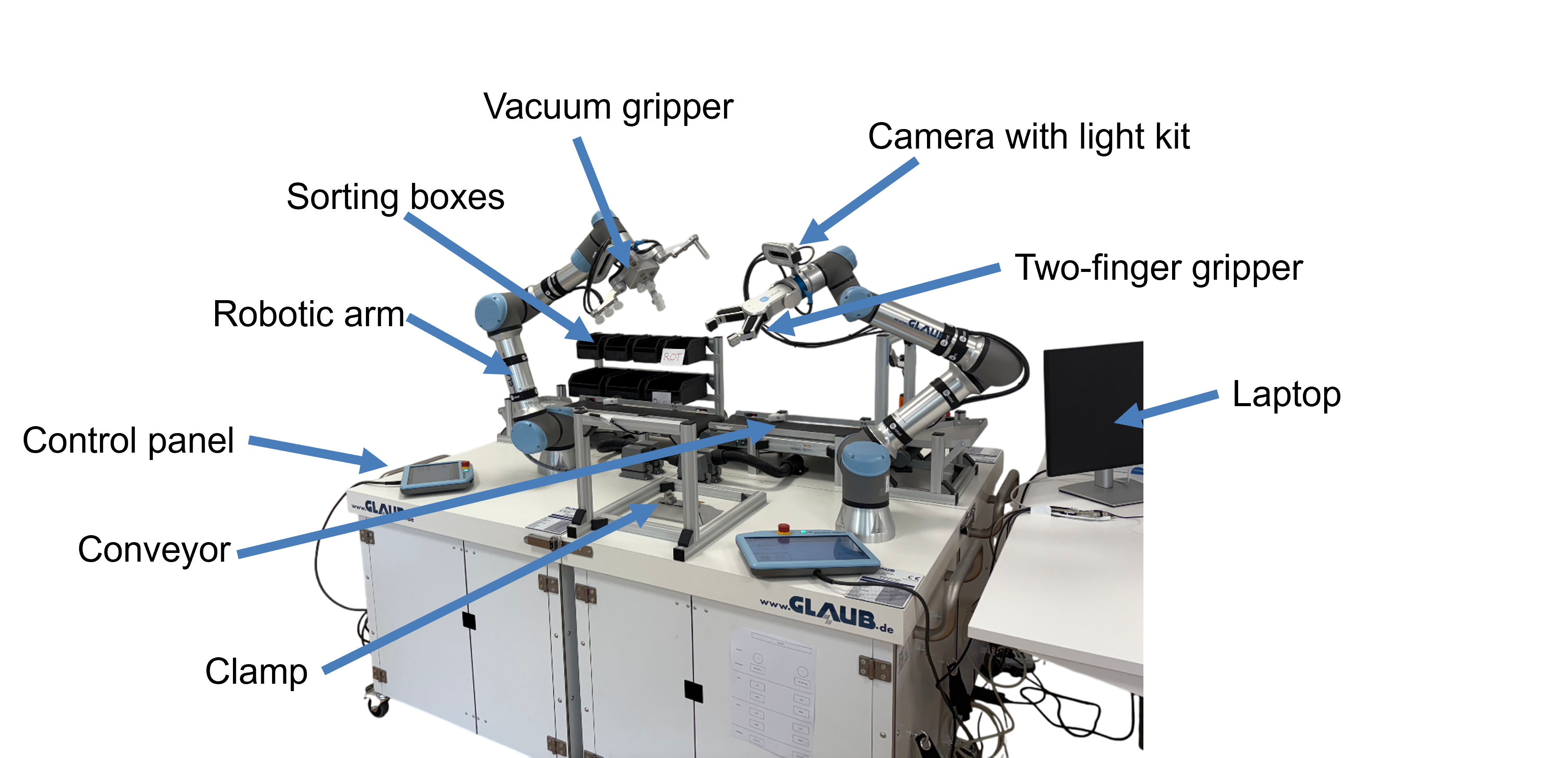}
        \caption{All important components of the robotic station}
        \label{fig:system}
    \end{minipage}
\end{figure}

The main task of the project is, therefore, the automated disassembly of brick assemblies, such as depicted in \autoref{fig:brick_construction}, which consist of 4x2, 2x2, and 2x2 round blocks of different colors. 
The main goal is to enable the system depicted in \autoref{fig:system} to disassemble the different brick assemblies. 

These brick assemblies are configured freely in compliance with specific rules, which ensure that the hardware effectors are capable of picking the respective brick. This means that the system must behave adaptively to ensure the proper disassembly. To enable such a system, the students focus on the following core contributions:

\begin{enumerate}
    \item The training and implementation of the computer vision component, responsible for capturing the brick constructions and analyzing their composition.
    \item The conception and implementation of the planning component to derive the disassembly sequence for the actuators. 
    \item The design and implementation of the actual controller ecosystem to manage the automation systems and robots.
\end{enumerate}

As a hardware platform, the students are using a disassembly station, consisting of two robots and two conveyor belts, as shown in \autoref{fig:system} \cite{10949991}. 
The robot system features a ROS backbone to control both the robots and the adjunct technical installations.
By default, the system features different topics for moving, rotating, grasping, controlling digital outputs, getting joint states, getting states of digital inputs, controlling the conveyor, and controlling the clamp.

The students started with an introduction session, onboarding them on robotics in general, as well as the functionalities of the robot system and the ROS drivers for controlling both robots.
Since the functionalities of some of the system's components have been implemented with ROS, a ROS bridge is required in order to integrate all functionalities into the new ROS2 system that the students are implementing to achieve the project goal.
Therefore, one of the schools at the beginning of the project offered a deep Dive into ROS \& ROS2 as well as information on the hardware system itself. 

After the initial onboarding of the students was completed, the students applied the knowledge about the robot functionalities to approach their first sprint goals: Determine the position of the robot to streamline the start and end sequence of the disassembly process. To solve the first assignment, the students had to put their knowledge to the test on how to read out the tool joint coordinates of the robot. This first task was done during the first two sprints, which were mentioned in \autoref{fig:Scrum}.

The programming of both the start and the end sequences in their corresponding ROS nodes was the next consecutive task and, therefore, the goal of the next sprint for the student team. The team, therefore, had to integrate previously defined ROS topics into new nodes, applying their understanding of the ROS framework. 
These nodes contain the most crucial topics for moving the robots, grasping objects with the gripper, activating the conveyor belts, as well as the triggers for the photoelectric sensors. In addition, the nodes should be callable via ROS commands in order to embed them into a function system for automated disassembly.
The tasks, therefore, offered the students to extend their knowledge gained during the schools by taking design considerations for the overall ecosystem into account, which took place while sprint three was ongoing.

The overall conception of the ROS ecosystem for the disassembly system aimed to test the developed understanding and further advance the project. 
This was the goal of the fourth sprint.
The previously designed node had, therefore, needed to be integrated into a larger system. This involved the integration of the component detection node for detecting the different components of the brick assemblies and the decision-making node for selecting the appropriate next brick. 

After setting up the overall robot system architecture and the hierarchical view, the team focused on improving some of the previously implemented functionalities. Therefore, the control of both robots was changed from a joint angle-based control to Cartesian coordinate navigation. The challenge was to draw the relationship between the robot's base frame and the corresponding coordinates and the bricks detected by the AI. The coordinate system of the AI detecting the brick and the world frame needed to be aligned to enable the precise grasping of the bricks in the disassembly process. This included, as well, the consideration of the location of the robot's camera in relation to the bricks. The relation between the two already connected frames can be seen by the ROS topics from the ROS node "/tf2". This is a complex goal, which requires multiple tasks and the connection between the robotic part and the AI part. For that reason, these goals were tried to be accomplished by the fifth and sixth sprints.

The final tasks and last sprint goal in the project were the integration of the system and the completion of the documentation of the project. For the final integration, the students had to implement all the necessary components of the system and had to test them on a disassembly cycle. The documentation of the project finally benefited as well from the previously learned techniques during the school sessions. This is finalized by the end of the project, sprint seven, as visible in \autoref{fig:Scrum}.

\section{Conclusion} \label{conclusion}
The paper introduces the application of the agile project management framework Scrum for projects in the domain of cyber-physical systems. Within the scope of the paper, an extension of the original Scrum-based concept is proposed to ensure that students exhibit the necessary knowledge in automation and robotics to contribute to the final goal while already participating in the main project tasks. Due to the combination of a preliminary, generalized project in the second semester with the subsequent three domain-related projects, as described in \autoref{Educational Approach}, students are enabled to enhance their knowledge throughout the program. The applicability of the concept is demonstrated through an exemplary project as shown in \autoref{projectintro}. Future research should focus on the approach's transferability to other domains and on optimizing the balance between school-related teaching sessions and sessions focused on the technical project goals. Especially, this direction is highly valuable, since the study program itself does require constant assessment regarding the adaptation of the schools based on the project content. 

\section{Acknowledgment}
The authors would like to thank all the students who participated in this project. 
This work has been developed in the project “Decentralized and Intelligent Circular Economy” (Research Grant Number 5603 1050) and is funded by the state government of Lower Saxony, Germany.

\printbibliography
\end{document}